\documentclass{article}
\usepackage{ijcai24}

\usepackage{times}
\usepackage{soul}
\usepackage{url}
\usepackage[hidelinks]{hyperref}
\usepackage[utf8]{inputenc}
\usepackage[small]{caption}
\usepackage{graphicx}
\usepackage{amsmath}
\usepackage{amsthm}
\usepackage{booktabs}
\usepackage{algorithm}
\usepackage{algorithmic}
\usepackage[switch]{lineno}

\title{Proposal Report for the 2nd SciCAP Competition 2024}

\author{
    Pengpeng Li$^1$, Tingmin Li$^1$, Jingyuan Wang$^1$, Boyuan Wang$^1$, Yang Yang$^1$\thanks{Corresponding author:Yang Yang(yyang@njust.edu.cn)}
    \affiliations
    Nanjing University of Science and Technology
}

\begin{document}

\maketitle

\begin{abstract}
   In this paper, we propose a method for document summarization using auxiliary information. This approach effectively summarizes descriptions related to specific images, tables, and appendices within lengthy texts. Our experiments demonstrate that leveraging high-quality OCR data and initially extracted information from the original text enables efficient summarization of the content related to described objects. Based on these findings, we enhanced popular text generation model models by incorporating additional auxiliary branches to improve summarization performance. Our method achieved top scores of 4.33 and 4.66 in the long caption and short caption tracks, respectively, of the 2024 SciCAP competition, ranking highest in both categories.
\end{abstract}

\section{Introduction}

Tables, images, and other objects within a paper can help readers quickly grasp the main ideas. Therefore, extracting descriptive information related to these objects from the text has garnered increasing attention\cite{multimodal_1} \cite{yang2021rethinking}. With the development of the transformer\cite{transformer} architecture, many transformer-based methods have been applied to the field of text generation, efficiently performing tasks such as text segmentation, summarization, and description generation\cite{luddecke2022image}\cite{yang2019semizhan}. Extracting descriptive information for related objects has also gained significant focus\cite{yang2022exploiting}. Compared to summarizing the document itself, summarizing descriptions related to objects is more challenging. This task requires attention not only to the text's context but also to the relevance between the objects and certain parts of the text. However, current methods often treat this task as a simple text generation problem.

Mainstream text generation methods include generating summaries from plain text and using multimodal techniques to align text and image information to generate summaries\cite{wangfengqiang}\cite{yang2022domfn}. For the former, a common approach is to perform basic OCR on the objects or use models like CLIP\cite{clip} to generate object descriptions. Most of these methods focus only on the information contained in the original or related text, without fully utilizing the information about the objects themselves. For multimodal methods, large language models such as LLaMA or GPT\cite{gpt} and multimodal models such as OFA\cite{ofa} and BLIP2\cite{blip} are widely used. While these models achieve high accuracy, they have limitations and can serve as auxiliary tools in the model fusion stage. Furthermore, previous methods typically rely on a single inference pass to obtain results, which can be biased. Lengthy descriptions often contain substantial background information that acts as noise, affecting the model's performance\cite{fu2024noise}.

Based on the above incomplete reasoning process, we propose our own summarization model. To address the incomplete utilization of information, we use PaddleOCR, an OCR recognition tool from the open-source Paddle model, to obtain high-quality OCR information about the objects. For the inference process, we deviate from the single-pass inference method. We designed a filtering method that divides paragraphs into multiple chunks, allowing the model to identify chunks that are useful for generating the correct answer. After an initial extraction of lengthy text, we perform secondary inference on the descriptions related to the objects. This segmented inference approach allows the model to focus on information with strong and weak relevance to the objects separately, reducing the impact of irrelevant text on the results.

Our contributions can be summarized as follows:

1. We utilize information stored within the objects themselves, which is often overlooked, as input for descriptive information. This allows the model to focus more on the descriptions related to the relevant objects rather than the text itself.

2. Using a two-stage training and inference process, our model minimizes attention to text weakly related to the objects while reducing the noise interference from background information.

\section{Related Work}
The task of generating descriptions based on object or text information has seen significant development\cite{caption_1}\cite{caption_2}\cite{caption_3}. Current mainstream methods are broadly categorized into pure text generation and multi-modal generation.

Text Generation. This method focuses primarily on extensive descriptive information about the objects. Models excel at extracting accurate descriptions from textual information. For non-textual information, such as images and tables, models like CLIP\cite{clip} are often used to generate a rough description of the object, or OCR technology is employed to recognize the characters and lines inherent to the object. These pieces of information are then input into the model along with the text for supervised learning. While this approach is easy to implement, it often fails to yield satisfactory results due to the limited learning capacity of text generation models. Additionally, the fusion of object information and textual information is not well-integrated.

Multi-Modal Generation. Multi-modal generation involves using two types of modalities at the input stage: one directly encodes the object, and the other encodes the text information\cite{yang2019semi}\cite{multimodal_2}. Thanks to the rapid advancements in technologies like LLaMA and GPT\cite{gpt}, the alignment issues between modalities have become negligible. Consequently, using large multi-modal models generally produces excellent results. However, due to the enormous parameters and technical complexity of these models, making technical improvements is not straightforward.

Overall, improving text generation methods is both practical and efficient. We have made enhancements to the text generation model to fully utilize known information, enabling it to understand and efficiently extract all relevant information. Detailed descriptions of these methods can be found in Chapter 3.
\begin{figure*}[!h]
	\centering
	\includegraphics[width=\linewidth]{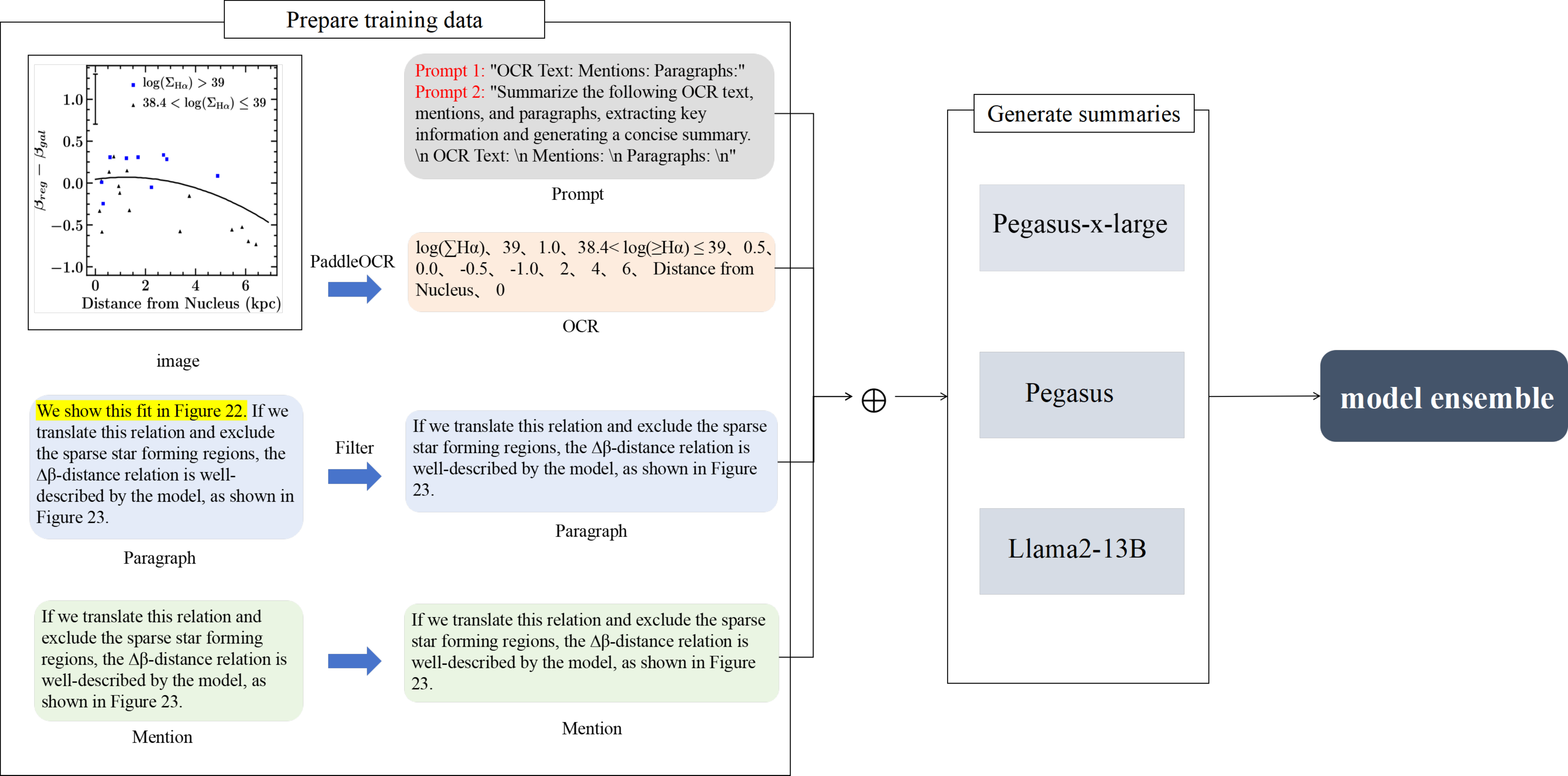}
	\caption{Overall Architecture. Our solution consists of three main stages, which includes prepare training data, generate summaries and Model-ensemble. }\label{fig:overview}
\end{figure*}

\section{Methodology}
\subsection{Overall Architecture}

Figure \ref{fig:overview} illustrates the overall architecture of our solution, which contains three components: Prepare training data, Generate summaries and Model-ensemble. Next, we will specifically introduce the three components mentioned above.

\subsection{Prepare training data}

\begin{figure}[t]
	\centering
	\includegraphics[width=\linewidth]{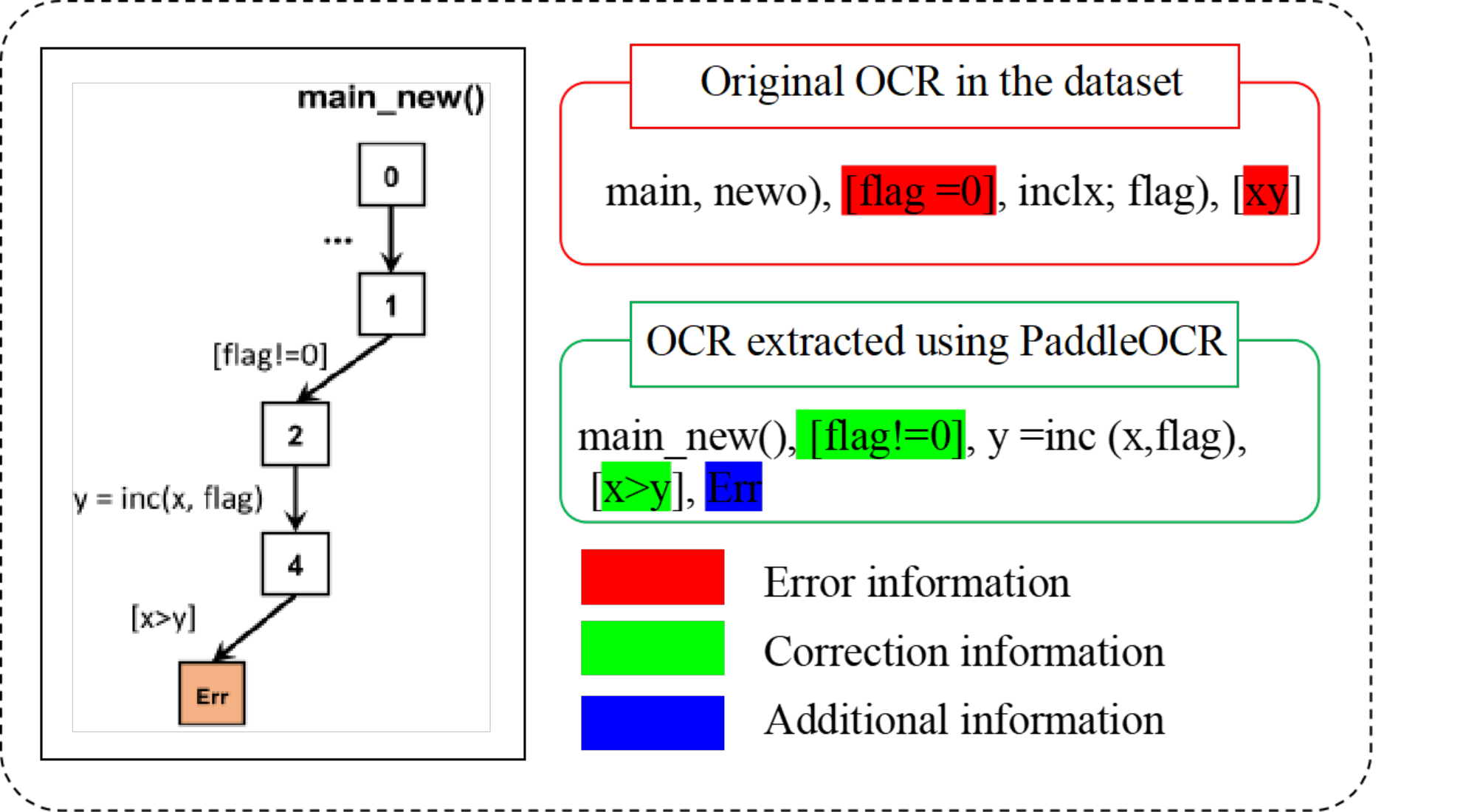}
	\caption{Comparing OCR results from the original data with those extracted using PaddleOCR, the original OCR contains error information. while OCR captured by PaddleOCR corrects these errors and includes information missed in the original data. }\label{fig:ocr}
\end{figure}

\textbf{OCR information extraction from images} aims to convert textual information contained within images into a text format, enabling the generation of more accurate and detailed image descriptions, thereby enhancing understanding of the image content. However, we found that OCR information in the images provided by the official dataset is not accurate, as illustrated in the figure \ref{fig:ocr}, making it difficult to generate precise image descriptions. To address this issue, we have chosen to use the PaddleOCR to re-extract OCR information from the images. This step aims to ensure that we obtain more accurate and reliable textual data to support subsequent summary generation tasks. 

\textbf{Filter gold sentence in paragraphs} aims to filter out redundant information in paragraphs to create more refined and effective training data, which is beneficial for model training. Specifically, we first concatenate the provided multiple paragraphs into a complete paragraph, then use the SpacyTextSplitter tool to divide this complete paragraph into several different blocks, represented by \(c\). Afterwards, we construct two types of input sequences: one is the mention \(m\) combined with the chunk \(c\), and the other consists only of mention \(m\), which can be represented as \(m \oplus c\) and \(m\) respectively. We will measure the differences in the probabilities of model \(M_\text{filter}\) to generate the given output \(o\), this process can be represented as \[I(c) = \frac{M_{\text{filter}}(o \mid c \oplus m)}{M_{\text{filter}}(o \mid m)}\]. Furthermore, to achieve the goal of filtering out irrelevant sentences, we set a threshold \(\lambda\). Sentences with probabilities below this threshold are filtered out, thereby the set of sentences that ultimately make up the paragraph is \(\{ c \mid I(c) \geq \lambda \}\) , considering that the sentences in the set are beneficial for generating the final summary. Finally, we concatenate all the sentences in the set to form the final paragraph, denoted as \(\hat{p}\).

However, during the training phase, since the correct answers are already known, we can filter out irrelevant information from the paragraph using the aforementioned method. But during the testing phase, as the correct answers are unknown, in order to filter out irrelevant information from the test data, we trained the flan-t5-xl model. In the construction of training data for the model, the input consists of the prompt "Filter out the irrelevant information in the text:" concatenated with the original paragraph \( p \), and the model's output is \(\hat{p}\) obtained through the above method. The training process can be represented as \( M_{filter}(\hat{p} \mid \text{prompt} \oplus p) \). In the testing phase, given the paragraph \(p\) and the prompt, we use the model \(M_{filter}\) to predict the filtered text, \(\hat{p} = M_{filter}(\text{prompt} \oplus p)\). The filtered text is then used in the summary generation phase. Through this method, we filter out paragraph descriptions that are beneficial for generating image captions and reduce the model's overhead during the summary generation stage.

\textbf{Construction of prompts} aims to ensures clarity in instruction and consistency in data input across different model architectures, facilitating effective training and evaluation. We further develop the training dataset for use in the summary generation phase. We have designed two types of prompts for the Pegasus\cite{pegasus}, Pegasus-x-Large, and LLaMA2-13B models, respectively. These are structured as follows:

Prompt 1: "OCR Text: Mentions: Paragraphs:"

Prompt 2: "Summarize the following OCR text, mentions, and paragraphs, extracting key information and generating a concise summary. OCR Text: Mentions: Paragraphs: "

\subsection{Generate summaries} Through the steps outlined in the previous section, we can obtain the text description \( t \) corresponding to each image. During the training phase, the input to the generative model \( M_{gen} \) is the constructed description \( t \), and the ground truth is the actual caption \(ans \) corresponding to the image, used as the output, which can be formalized as \( M_{gen}(\text{ans} | t) \). In the inference phase, we input the constructed text description \( t \) into the model, enabling the model to generate a summary for the given text.

\begin{table}[h]
\centering
\begin{tabular}{lccccc}
\hline
Method     & Blue4 & R-1  & R-2  & R-1-n & R-2-n \\
\hline
Base       & 0.13  & 0.40 & 0.25 & 2.38  & 3.81 \\
+PaddleOCR  & 0.11  & 0.37 & 0.28 & 2.48  & 4.02 \\
+LLaMA    & 0.16   & 0.43 & 0.27 & 2.56  & 4.05   \\
+Filter      & 0.08  & 0.40 & 0.24 & 2.23  & 4.11  \\
combine    & 0.06  & 0.39 & 0.24 & 2.37  & 4.66  \\
\hline
\end{tabular}
\caption{Optimization and results  on the short caption.}
\end{table}

\subsection{Model-ensemble}
 Model ensemble is the final stage. Specifically, we train the Pegasus, Pegasus-x-Large, and LLaMA2-13B models and obtain their weights at different epochs. For Pegasus and Pegasus-X-Large, we use the weights from the 4th, 5th, and 6th epochs for inference, and each model generates 16 candidate summaries. For the LLaMA2-13B model, we use the weights from the 3rd, 4th, 5th, and 6th epochs for inference, with each weight generating one candidate summary. Thus, for each image, we obtain 100 candidate summaries, from which we select the summary that best describes the image. We adopt a method of sequentially assuming each of the 100 candidate summaries as the correct summary. The score of this summary is the average of the cumulative scores of the ROUGE-2-normalized scores compared to the other summaries, which can be formulated as:
 \[
s_n = \frac{1}{|R_N| - 1} \sum_{r_m \in R_N \setminus \{r_n\}} \text{sim}(r_n, r_m)
\], where \(s_n\) is defined as an averaged score to all the other captions \((r_m \in R_N \setminus \{r_n\})\), \(\text{sim}(r_n, r_m)\) is the ROUGE-2-normalized score between two captions \(r_n\) and \(r_m\) , and \(R_n\) represents all the candidate summaries. Ultimately, we select the summary with the highest score as the caption for the image.

\section{Experiment}

\subsection{Dataset}
The dataset used in this study is the official dataset provided, which contains approximately 500,000 data samples. Each sample includes a description of an object, a paragraph of original text about the object (without preliminary extraction), and several sentences related to the object (extracted from the paragraph). About 400,000 samples were used for training, and 47,639 samples were used for testing.

\subsection{Details}
We used two A100 GPUs to train the base models, including pegasus-x-large, and Pegasus, among others. The training was conducted for varying epochs, ranging from 3 to 10, with a learning rate set at 1e-5.

\subsection{Results}
Our experimental results for the short track is presented in Tables 1. BLEU (Bilingual Evaluation Understudy) is a metric used to evaluate the quality of machine-translated text against one or more reference translations by measuring the n-gram overlap. ROUGE-1 (Recall-Oriented Understudy for Gisting Evaluation) calculates the overlap of unigrams (single words) between the generated text and reference text, focusing on recall. ROUGE-2 extends this by calculating the overlap of bigrams (two-word sequences), providing a more detailed measure of similarity. ROUGE-1 Normalized adjusts the ROUGE-1 score by the length of the reference text, giving a normalized recall value to account for text length differences. ROUGE-2 Normalized does the same for ROUGE-2, normalizing the bigram overlap score to ensure comparability across texts of varying lengths.

\section{Conclusion}
This paper outlines the methods we employed in the 2024 2nd SciCap competition. We improved upon traditional text generation models by addressing the limitations in extracting information from extensive text and the underutilization of object-specific information. Our approach maximizes the extraction of information from both the text and the objects themselves. Under this method, standard text generation models outperformed large language models in summarizing descriptive text.

\bibliographystyle{named}
\bibliography{ijcai24}

\end{document}